\pdfoutput=1
\documentclass[preprint]{vgtc}
\makeatletter
\renewcommand{\ps@plain}{%
  \renewcommand{\@oddhead}{\hfil\textit{\small Preprint. Under review for PacificVis 2027.}\hfil}%
  \renewcommand{\@evenhead}{\@oddhead}%
  \renewcommand{\@oddfoot}{\footnotesize Repository Link: \url{https://github.com/Ni2002ka/residual_stream_vis/tree/v1.0-pacificvis}\hfil\textrm{\thepage}}%
  \renewcommand{\@evenfoot}{\@oddfoot}}
\let\ps@empty\ps@plain 
\makeatother
\usepackage{times}

\usepackage{mathptmx}
\usepackage{booktabs}
\usepackage{xcolor}
\usepackage{amsmath}
\usepackage{float}
\usepackage{stfloats}
\usepackage{enumitem}

\newcommand{\LL}{logit lens}
\newcommand{\JL}{J-lens}

\onlineid{1217}
\vgtccategory{System}
\vgtcinsertpkg

\title{MicroLens: Read, Trace, and Rewire the Residual Stream with Two Lenses}

\author{Nika Zahedi\textsuperscript{1} \and Fangyi Zhu\textsuperscript{2} \and Anna Wu\textsuperscript{1}}
\affiliation{\scriptsize \textsuperscript{1}Stanford University \quad \textsuperscript{2}Independent researcher}

\teaser{
  \centering
  \includegraphics[width=\linewidth]{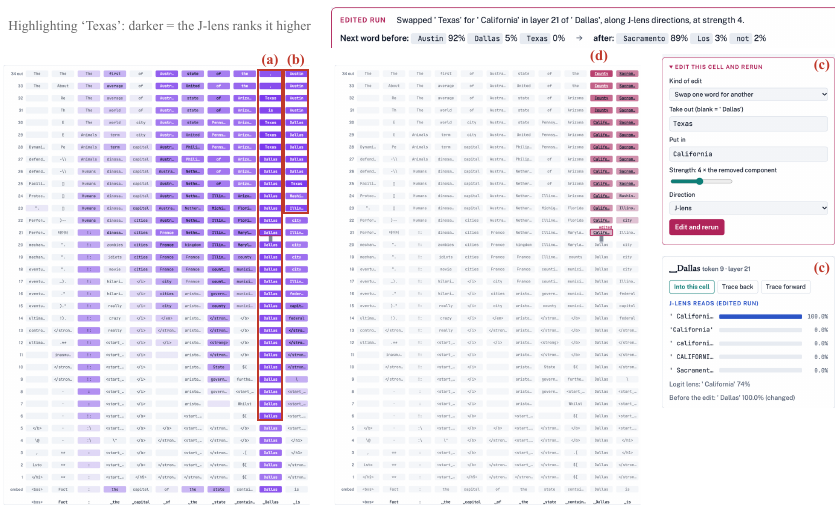}
  \caption{MicroLens on a two-hop question (Gemma-3-4B-it, ``Fact: the capital of the state containing Dallas is'').
    Each cell is one token's residual-stream vector after one layer, labeled with the \JL{}'s top word. Left:
    highlighting \emph{Texas} shades each cell by how highly the \JL{} ranks it. (a)~The \texttt{Dallas} cells rank
    \emph{Texas} highly from layer~6 while their top word is still \emph{Dallas}, and make it the top word from
    layer~28. (b)~The last column reads \emph{Texas} at layer~25, then \emph{Austin}. (c)~Swapping \emph{Texas} for
    \emph{California} at \texttt{Dallas}, layer~21 (\JL{} directions, strength~4) changes the answer from
    \emph{Austin} (92\%) to \emph{Sacramento} (89\%). (d)~``What changed'' shading: due to causality, only cells above and to the right
    of the edit move. The readings in (a) and (b) suggest that the model works out \emph{Texas} on the way to \emph{Austin}. The edit in
(c) and (d) lets us test that it uses \emph{Texas}, since changing it changes the answer.}
  \label{fig:teaser}
}
\abstract{
Lenses that decode hidden states into words are a popular way to look inside language models, but a
heatmap of decoded words only shows what a vector reads as and cannot distinguish whether, or how much, the model uses the information. We present
an interactive viewer that puts reading, tracing and editing on the same layers-by-tokens grid of the residual stream.
In the viewer, every cell can be decoded with two lenses, the \LL~\cite{nostalgebraist2020} and the Jacobian lens
(\JL)~\cite{workspace2026}, traced back to the tokens it drew from, and edited, with the rest of the model rerun on
the spot. We demonstrate three usage scenarios on Gemma-3-4B-it, backed by sweeps over a few hundred prompts, to show what this
tool can discover. First, the two lenses give different directions to write with, each suiting a different edit:
the \LL's word direction sets what a token outputs, while the \JL's direction changes what later tokens read.
Second, in two-hop trivia questions (the capital of the state or country containing a given city), the state or
country is readable at the city token from as early as layer~7, and swapping it there can redirect the answer
92--98\% of the time; however, at the answer position, the state or country information is not used. Third, when a chat model answers questions with a step-by-step reasoning prompt, the correct answer is often already the top word in the residual stream before any reasoning is written---specifically, for 89--92\% of retrieval questions and 39--63\% of multi-hop and serial ones, far above their respective distractors ($\leq$5\%). However, ordering puzzles are an exception, as the model works them out only once it writes its final reasoning steps. Furthermore, for answers that are visible in the residual stream, the \JL{} reads them a layer or two earlier than the \LL{} across the different types of question.
}


\begin{document}

\firstsection{Introduction}
\maketitle

A transformer language model's \emph{residual stream} holds one vector for each token of its input. Every layer
reads these vectors and adds its output to them. Gemma-3-4B-it~\cite{gemma3}, for example, has 34 layers and
2,560-dimensional vectors. Recording every token's vector after every layer gives a grid: one column per input
token, one row per layer, from the input embeddings at the bottom to the last layer at the top. At the top, the model
turns each vector into a probability distribution over its vocabulary, using the resulting distribution to predict the next token.

A \emph{lens} applies this readout to any cell of the grid, giving the distribution over words the model would
predict if it stopped at that layer. The \LL~\cite{nostalgebraist2020} uses the model's own final norm and
unembedding matrix unchanged. More refined lenses first map the vector from its layer toward the top, with a learned
map (the tuned lens~\cite{belrose2023tuned}) or the average Jacobian of the layers above (the Jacobian lens, or
\JL~\cite{workspace2026}), and often read an idea several layers earlier. Labeling each cell with its most probable
word gives a lens heatmap.

Such a heatmap invites two mistakes. A word that is readable in a cell may play no part in the model's computation,
and a word that matters may not be readable. Causal methods can be used to test whether (and how much) a cell matters. Examples are activation
patching~\cite{meng2022rome, zhang2024patching}, steering~\cite{turner2023actadd} and attribution
graphs~\cite{ameisen2025circuit, lindsey2025biology}. However, these methods are usually run as scripts, separate from
the intuition that a lens picture gives. Through making the flow of information within the model visible and easily
editable, we aim to bridge the gap between that intuition and the causal tests that confirm it.

We present \textbf{MicroLens}, a viewer that puts reading, tracing and editing on the same grid (\cref{fig:teaser}). A user
reads any cell with either lens, traces where its information came from, edits it, and sees every layer above
recompute. We also present three usage scenarios on Gemma-3-4B-it, each motivated by a pattern we first noticed in the viewer
and then confirmed by a sweep with controls. First, the two lenses give different directions to edit with
(\cref{sec:directions}). Second, in two-hop questions, the intermediate entity can be readable at several tokens, but the model uses it at
only one of them (\cref{sec:twohop}). Third, when the model reasons step by step, the answer is often readable
before the reasoning begins (\cref{sec:cot}). Our contributions:
\begin{itemize}
  \item An open-source, local viewer for Hugging Face decoder models that combines two lenses, attention-flow
        tracing, four kinds of cell edits and chain-of-thought tracking on one grid (\cref{sec:tool}).
  \item Findings that guide edit design: Different lenses provide edit directions that are useful depending on what the edit is for: To set
    what the final token \emph{outputs}, use the \LL's direction. To change what \emph{later tokens read}, use the
    \JL's, which works only where the \JL{} can read the word (\cref{sec:directions}).
    \item In two-hop recall, the intermediate entity is readable at two tokens, but editing shows that the model uses
        it at only one of them, which a lens reading alone (as in previous lens tools) cannot reveal (\cref{sec:twohop}).
    \item In step-by-step reasoning, the answer is often readable before the reasoning tokens begin, and the \JL{} reads
        it earlier than the \LL{} (\cref{sec:cot}).
\end{itemize}

\section{Related Work}
\;\;
\textbf{Lenses:} The \LL~\cite{nostalgebraist2020} applies the model's final norm and unembedding to an intermediate
vector. The tuned lens~\cite{belrose2023tuned} learns an affine map per layer; the \JL~\cite{workspace2026} instead
uses the average Jacobian from a layer to the top, estimated once on web text. Patchscopes~\cite{ghandeharioun2024patchscopes}
decode a vector by patching it into another prompt. Lens readouts have been used to argue that multilingual models
pass through English~\cite{wendler2024llamas} and that models resolve multi-hop questions
internally~\cite{yang2024multihop, biran2024hopping}.

\textbf{Causal analysis:} Activation patching replaces a cell with its value from another prompt and measures the
effect~\cite{meng2022rome, zhang2024patching, heimersheim2024patching}. Steering adds a direction~\cite{turner2023actadd,
todd2024function}. Geva et al.~\cite{geva2023dissecting} show that attention moves subject attributes to the last
token at specific layers, which we observe directly as a ``handoff'' (\cref{sec:twohop}). Attribution
graphs~\cite{ameisen2025circuit, lindsey2025biology} build feature-level causal graphs. We borrow their two-hop case
study and test it with lens directions on an open model.

\textbf{Visual tools:} Attention visualizers~\cite{vig2019bertviz}, LM-Debugger~\cite{geva2022lmdebugger}, the
LLM Transparency Tool~\cite{tufanov2024lmtt} and Neuronpedia~\cite{neuronpedia} expose attention, neurons, features or
information flow. Surveys cover visual analytics for deep learning broadly~\cite{hohman2019survey}. Our viewer is
narrower: it is built around the residual-stream grid and makes every cell both readable with two lenses and
editable, so that a reading can be tested and traced.

\section{The Viewer}
\label{sec:tool}
  \;\;
\textbf{Grid:} A prompt runs once with hooks that record every block's input and output, the attention and MLP
writes, and per-head attention weights. Columns are tokens and rows are layers, from the input embedding (row~0) to
the output (top). Each cell shows the top word under the selected lens; at the top both lenses give the model's
actual prediction. A user can switch lenses for the whole grid.

\textbf{Cell details:} Clicking a cell lists both lenses' top words and splits the block's update into what was kept
from below, what attention wrote and what the MLP wrote. It also shows which earlier tokens and heads the attention
write came from. Each source is weighted not by its raw attention weight, but by the size of the vector it adds to the
cell: the attention weight times the length of its value vector~\cite{kobayashi2020norm}. A token can receive a high
attention weight yet contribute little if its value vector is small. Chaining these shares across layers gives an
approximate trace back to the input or forward to the output, a norm-weighted rollout~\cite{abnar2020flow}.

\begin{figure}[tb]
  \centering
  \includegraphics[width=\columnwidth]{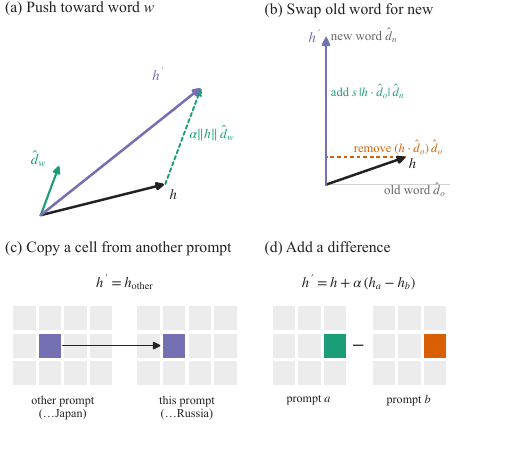}
  \caption{Schematic of the four edits to a cell $h$. (a)~Push adds a word's direction $\hat d_w$, scaled by the
    cell's own length. (b)~Swap removes the cell's component along the old word and adds $s$ times its size along
    the new word, leaving the rest of the cell unchanged. (c)~Copy replaces the whole cell with the same cell from
    another prompt. (d)~Add a difference adds the difference between two prompts' cells.}
  \label{fig:edits}
\end{figure}

\textbf{Edits:} A user can change the selected cell and rerun every layer above it, comparing the grid before and
after. The ``what changed'' shading shows how far the edit spreads (due to causality, only cells above and to the right can move). We denote
$h$ for the selected cell, the residual-stream vector at one layer and token, and $h'$ for its edited value. Four edits
are available (\cref{fig:edits}):
\begin{itemize}
    \item \emph{Push} moves the cell toward a word $w$:
\begin{equation}
  h' = h + \alpha\,\lVert h\rVert\,\hat d_w ,
\end{equation}
where $\hat d_w$ is $w$'s direction under the chosen lens and $\alpha$ sets the size of the
push as a fraction of the cell's own length $\lVert h\rVert$; a negative $\alpha$ pushes away from $w$.

\item \emph{Swap} replaces an old word $o$ with a new word $n$:
\begin{equation}
  h' = h - (h\cdot\hat d_o)\,\hat d_o + s\,\lvert h\cdot\hat d_o\rvert\,\hat d_n .
\end{equation}
The dot product $h\cdot\hat d_o$ measures how much of the old word the cell holds. The second term removes that
amount, and the third adds $s$ times as much along the new word, where the \textbf{strength} $s$ sets how strongly the new word
replaces the old. The rest of the cell is left unchanged.

\item \emph{Copy from another prompt} replaces the whole cell with the cell at the same layer and token position of a different,
counterfactual prompt, $h' = h_{\text{other}}$. This is the same as activation patching~\cite{meng2022rome, zhang2024patching, heimersheim2024patching}. For example, the cell for
\texttt{Russia} can be copied from the same prompt with \texttt{Japan}. A swap changes only the part of the cell that
reads as the old word. A copy brings in everything the model computed at that token from the other prompt, including
what it gathered from earlier tokens through attention, so it serves as a ceiling for what a swap can achieve. Note that a swap needs only the two words, whose directions come from a lens, while a copy needs a second prompt. The second prompt is the original with only the entity replaced (e.g.\ \emph{Japan} for \emph{France}), so the two
prompts line up token by token, and the copy takes the cell at the same layer and position.

\item \emph{Add a difference} adds the difference between the last-token cells of two prompts $a$ and $b$: a steering
vector~\cite{turner2023actadd}, $h' = h + \alpha\,(h_a - h_b)$. This operation needs no lens: the difference itself defines the direction.

\end{itemize}

Push and swap need a direction $\hat d_w$ for each word, and the two lenses give two kinds. The \LL{} direction is the
word's row $u_w$ of the unembedding matrix, the vector the output layer uses to score $w$. The \JL{} direction is
$J_\ell^{\top} u_w$, scaled to unit length. Here $J_\ell$ is the layer's Jacobian: a matrix that maps a small change
to a cell at layer~$\ell$ to the change it causes at the output, averaged over many prompts. Multiplying $u_w$ by its
transpose transforms the word's output direction back into layer~$\ell$: it gives the change at layer~$\ell$ that,
passed through the layers above, most increases $w$'s score. \Cref{sec:directions} shows why both kinds are needed.

\textbf{Chain of thought:} For chat models, the viewer generates a response with the chat template and covers the
prompt and response in one grid. The ``\emph{Track a word}'' functionality shades every cell by a lens's rank for one word, so a user can
see whether an answer is present before the model writes its reasoning.

\textbf{Implementation:} The viewer is an open-source Python package that serves a local web page. It is built on Hugging Face Transformers~\cite{wolf2020transformers} and finds a model's transformer blocks automatically, so it works with GPT-2, Llama, Mistral, Qwen, Gemma, Phi, GPT-NeoX and most other decoder-only models. It loads the published \JL{}es for GPT-2 and every size of Gemma~3 (270M to 27B, base and instruction-tuned), which were fit on WikiText~\cite{neuronpedia}. It can also fit a new lens for any other model. On a laptop (Apple M2 Pro, 32\,GB), running Gemma-3-4B-it with its \JL{} takes 1--4\,s per prompt, depending on length (12--49 tokens). An edit reruns every layer above the edited cell and takes about 4--6\,s. The code is available at \url{https://github.com/Ni2002ka/residual_stream_vis/tree/v1.0-pacificvis}.

All experiments below use Gemma-3-4B-it~\cite{gemma3} with its published \JL, and only items the model answers
correctly whose words are single tokens. Intervals are bootstrap 95\% CIs; paired comparisons use exact McNemar tests.
Each scenario began in the viewer: a pattern we noticed on one prompt, such as a swap that worked on one antonym and
failed on another, or a state readable at a token whose edits did nothing. Because the viewer makes such patterns
quick to find and to test on a single prompt, we could turn each into a hypothesis and confirm it over hundreds of
prompts with the same code.

\section{Scenario A: Two Kinds of Edit Direction}
\label{sec:directions}

Take a five-shot prompt whose last lines are ``Q:~France / A:'', which the model completes with \emph{Paris}. To
make it answer \emph{Tokyo} instead, a user can edit the grid in two ways. One is to change the output directly:
\textbf{Push} the last token's cell toward \emph{Tokyo}. The other is to change what the model reasons from: \textbf{Swap}
\emph{France} for \emph{Japan} at the \texttt{France} token, and let the model work out \emph{Tokyo} itself. Exploring the viewer we found that each lens's direction suits
one of the above objectives better. For the push, the \LL's direction for \emph{Tokyo} works from early layers and the \JL's does not; for
the swap, the reverse holds. As a \textbf{\emph{ceiling}} for a best-case edit, we run the prompt with \emph{Japan} written in place of \emph{France}, take the cell at the
\texttt{Japan} token, and paste it over the \texttt{France} cell at the same layer, so that the cell is exactly what
the model holds for the new entity.

\textbf{Setup:} We tested this on four function-vector tasks~\cite{todd2024function} (country--capital,
product--company, antonym, English--French) as five-shot ``Q:~x / A:~y'' prompts, editing one cell at every layer. The \emph{push} adds a word's direction to the last
token's cell, scaled to the cell's own length ($\alpha=1$); it succeeds if the word, another item's answer, becomes the
top prediction. A \emph{swap} removes the entity token's component along the old entity's direction and adds twice
that amount along the new one's ($s=2$); it succeeds if the model gives the new entity's answer. Each edit uses either
the \LL{} direction (the word's row of the output matrix) or the \JL{} direction (that row times $J_\ell$), and a
random direction of the same size as the \JL{} edit serves as control. The ceiling for a swap is copying the whole
entity cell from a prompt that names the desired answer.

\subsection{Experiment A Results}
\begin{table}[tb]
\caption{Edits on Gemma-3-4B-it (34 layers), $n=25$--$100$ items per task. \emph{Push:} the first layer from which the
    pushed word is the top prediction for at least 90\% of items. \emph{Swap:} success at each direction's best layer,
    with that layer in parentheses. \emph{Random:} a control edit that changes the cell by the same amount as the \JL{}
    swap, but along a random direction. \emph{Copy cell:} success when the whole entity cell is copied from a prompt
    naming the new entity, our empirical ceiling for a single-cell edit.
    Random pushes succeed on at most 4\% of items at any layer. The rates are not sensitive to choosing the best layer
    on the same pairs it is scored on: choosing it on a random half and scoring the other half (20 split trials)
    changed them by at most 5 percentage points.}
  \label{tab:directions}
  \scriptsize
  \centering
  \setlength{\tabcolsep}{3pt}
  \resizebox{\linewidth}{!}{%
  \begin{tabular}{l|cc|ccc|c}
    \toprule
    & \multicolumn{2}{c|}{push: 90\% from layer} & \multicolumn{3}{c|}{swap along direction: success (layer)}
    & copy cell \\
    task & \LL{} & \JL{} & \LL{} & \JL{} & random & success \\
    \midrule
    \multicolumn{7}{l}{\textit{fact lookup}} \\
    country--capital & \textbf{7} & 28 & 6\% (1)  & \textbf{73\%} (20) & 0\% & 99\% \\
    product--company & \textbf{6} & 30 & 22\% (2) & \textbf{45\%} (21) & 1\% & 95\% \\
    \multicolumn{7}{l}{\textit{word}} \\
    antonym          & \textbf{7} & 22 & 6\% (1)  & 0\%                & 0\% & 80\% \\
    English--French  & \textbf{8} & 27 & 7\% (1)  & 1\% (20)           & 0\% & 88\% \\
    \bottomrule
  \end{tabular}}
\end{table}

\begin{figure*}[tb]
  \centering
  \includegraphics[width=\linewidth]{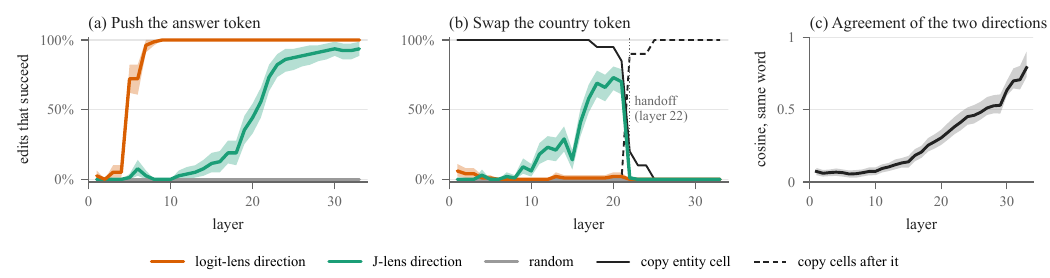}
  \caption{Each direction suits one edit (country--capital, Gemma-3-4B-it). (a) Pushing the answer token toward
    another country's capital: the \LL{} direction succeeds for 90\% of items from layer~7, the \JL{} direction only
    from layer~28 ($n=79$). (b) Swapping the country for another at its own token: the \JL{} direction succeeds in the
    middle layers ($n=100$ pairs), up to the handoff, where copying the cells after the country starts to beat copying
    the country's own cell (black lines, 20 pairs). Bands are bootstrap 95\% CIs. (c) Cosine between the two
    directions for the same word, over the 203 words pushed toward (band: interquartile range): nearly perpendicular
    through layer~12, converging near the top.}
  \label{fig:directions}
\end{figure*}
\;
\textbf{Push: use the \LL{} direction.} On every task, the \LL{} push makes the target the top prediction for at least
90\% of items from layer 6--8 on; the \JL{} push gets there only at layer 22--30 (\cref{tab:directions},
\cref{fig:directions}a). Compared layerwise, the gap is large: at the layer where the \LL{} push works best (7--9), it succeeds on every item, and the
\JL{} push on only 0--7\% of them ($p<10^{-7}$ on every task, exact McNemar test on the paired outcomes).

\begin{figure}[H]
  \centering
  \includegraphics[width=\columnwidth]{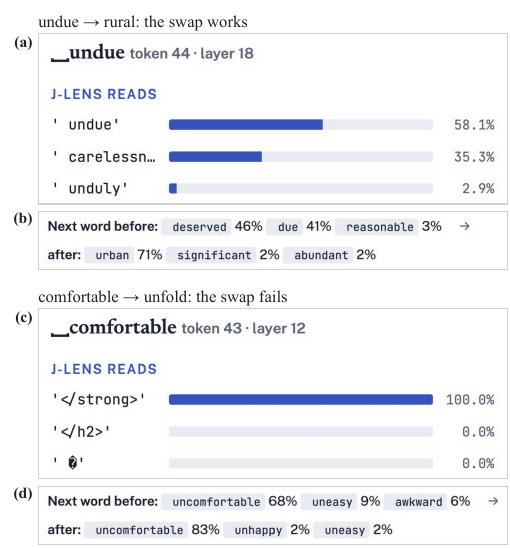}
  \caption{A \JL{} swap in the viewer (Gemma-3-4B-it, antonym prompts with five examples). (a)~At layer~18 the
    \JL{} reads the \texttt{undue} cell as \emph{undue} and related words, and (c)~swapping \emph{undue} for
    \emph{rural} there (strength~4) changes the answer to \emph{urban}. (b)~The \JL{} reads the
    \texttt{comfortable} cell as formatting tokens (layer~12 shown), and (d)~the same swap toward \emph{unfold}
    leaves the answer \emph{uncomfortable}. A \JL{} swap points the right way only when the \JL{} shows the word
    in its own cell; \emph{undue}$\to$\emph{rural} is the only antonym pair of 100 for which it succeeds.}
  \label{fig:viewerdemo}
\end{figure}

\textbf{Why the two differ:} For the same word, the two directions are almost uncorrelated through the middle of the
model: averaged over the 203 words pushed toward, their cosine is 0.06 at layer~8 and 0.31 at layer~20, and they
become similar only near the top (0.51 at layer~27, 0.79 at layer~33; \cref{fig:directions}c). They stand for
different things. The \LL{} direction is the word as the top layer reads it at this token: added to the last token,
it rides the residual connections to the output almost unchanged, and the model says the word.
The \JL{} direction is
what the layers above usually \emph{turn into} the word, which makes it an input to reason with rather than an
answer. At the last token there is nothing left to reason toward, so a \JL{} push fails until the top layers, where
the two directions coincide and it becomes, in effect, a \LL{} push. The converse also holds: the last token is the
wrong place to change what the model reasons with, as we will further see in (\cref{sec:twohop}).

\textbf{Swap: use the \JL{} direction, for facts.} On the two fact-lookup tasks, the \JL{} swap gives the new
entity's answer for 73\% and 45\% of pairs, at layers 20 and 21. At those layers the \LL{} swap succeeds on 2\% and
16\% ($p=1.5\times10^{-19}$ and $2.5\times10^{-5}$, exact McNemar), and a random edit of the same size on at most
1\%.

The ideal edit, our \emph{ceiling}, would paste in the cell the model computes for the new entity, taken from
the prompt with \emph{Japan} in place of \emph{France}; on a separate set of 20 pairs, the \JL{} swap achieves 79\%
and 62\% of what this paste does at the same layer. It falls short partly because it is small, moving only 1--5\% of
the cell's length. When it fails, the model keeps its original answer rather than giving a wrong one.

\textbf{Where the \JL{} cannot see the word:} On antonym and translation, both directions fail (at most 7\%), even
though pasting in the new word's cell succeeds 80--88\% of the time. The word is in the cell, but the \JL{} cannot
see it, and a \JL{} swap, which moves the cell from the old word's \JL{} direction toward the new word's, helps only
if it can. For names it can: at a country's own token the \JL{} ranks the country first or second among all words
from layer~1 on, and it sees the change the paste makes (across country pairs at layer~12, the new country is among
the five words the paste raises most 65\% of the time). For antonym and translation words it cannot: at layer~4 it
ranks the word only about 200th at its own token, and at layer~12 the new word ranks about 1{,}100th (antonym) and
2{,}100th (translation) among the words the paste raises, below formatting tokens (20 pairs per task). The two agree
pair by pair: across all 100 pairs, the worse the \JL{} ranks the old word at its own token, the less the swap's
change matches the paste's (Spearman $\rho=-0.62$).

Which words the \JL{} can see appears to depend on how many meanings they have: among antonym and translation words,
those with more WordNet senses rank worse at layer~4 ($\rho=0.45$ and $0.71$; \emph{flat} has 24 senses,
\emph{decorative} one). We read this as a consequence of averaging: $J_\ell$ is an average over contexts, so it keeps
what a word does consistently, which a name does and a word with many senses does not until context picks its sense,
around layer~16. This gives a check in the viewer before editing: if the \JL{} shows a word at its own token, its
direction can be used for a swap.

\textbf{Example:} A \JL{} swap of a word works only between the layer where the \JL{} first sees both words and the
handoff. In the antonym prompt ending in ``\texttt{Q: undue\textbackslash nA:}'' (answer \emph{due}), swapping
\emph{undue} for \emph{rural} (strength~4) gives \emph{urban}, but only at layers 18--20. At layer~12 the \JL{} reads
both cells as formatting tokens and the swap fails; at layer~18 it reads them as \emph{undue}, \emph{carelessness},
\emph{unduly} and \emph{countryside}, \emph{rural}, \emph{farmland}; from layer~21, the handoff, later tokens have
already read the word. For \emph{comfortable}$\to$\emph{unfold} the window is empty: the swap never gives
\emph{fold}, because the \JL{} reads both cells as formatting tokens (\texttt{</strong>}, \texttt{</h2>}) through
layer~12 and first ranks \emph{comfortable} in its top five at layer~23, after the handoff.

\begin{figure}[h]
  \centering
  \includegraphics[width=\columnwidth]{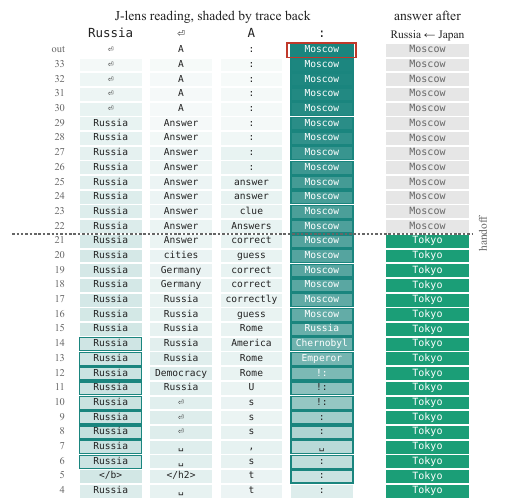}
  \caption{Finding the handoff in the viewer (Gemma-3-4B-it, ``\ldots Q: Russia\textbackslash nA:''). Left: the
    last four tokens, each cell labeled with the \JL{}'s top word and shaded by how much it feeds the output (Trace
    back from the outlined cell). Right: the answer after copying the \texttt{Russia} cell, at that layer, from the
    same prompt ending in \emph{Japan}. The copy changes the answer up to layer~21 and not after. Teal outlines: cells with trace-back weight of $>0.04$ to the output cell (red outline).}
  \label{fig:handoff}
\end{figure}

\textbf{The handoff:} Every swap stops working after layer 21--22: from that layer on, the later tokens already carry
the answer, and the entity cell no longer matters (\cref{fig:directions}b). The viewer shows both sides of this
(\cref{fig:handoff}). Tracing the output back from its cell follows the last token down and then into the entity
cell in the middle layers, and the copy edit finds the handoff with one click per layer: copying the \texttt{Russia}
cell from a prompt about Japan gives \emph{Tokyo} up to layer~21 and leaves \emph{Moscow} from layer~22. The handoff
bounds where any edit on the entity can work.

\section{Scenario B: Two-Hop Recall}
\label{sec:twohop}

Lindsey et al.~\cite{lindsey2025biology} find that Claude~3.5 Haiku answers ``Fact: the capital of the state
containing Dallas is'' through an internal \emph{Texas}, and that substituting \emph{California} yields
\emph{Sacramento}. We asked whether the viewer can find and test the same route in an open Gemma-3-4B-it model, with 46 U.S.\
city$\to$state$\to$capital and 51 world city$\to$country$\to$capital prompts.
The model answers 25 and 30 of them
correctly. Most of its errors stop halfway: the model answers with the state or country instead of its capital (\emph{Florida} for
Tampa, \emph{Switzerland} for Zurich), or repeats the city.

\begin{figure}[H]
  \centering
  \includegraphics[width=\columnwidth]{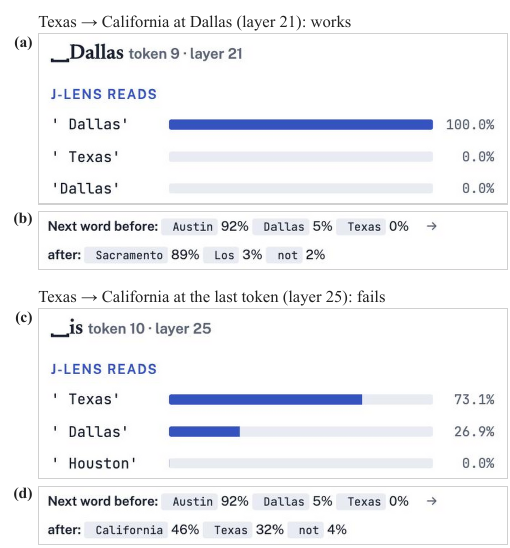}
  \caption{The Dallas question in the viewer (Gemma-3-4B-it). (a)~At \texttt{Dallas}, layer~21, the \JL{} reads
    \emph{Texas} second. (b)~Swapping \emph{Texas} for \emph{California} there (\JL{} directions, $s=4$) changes the
    answer from \emph{Austin} to \emph{Sacramento}. (c)~At the last token, layer~25, the \JL{} reads \emph{Texas}
    first. (d)~The same swap there makes the model name a state instead of reasoning on to its capital.}
  \label{fig:twohopviewer}
\end{figure}

\begin{figure}[H]
  \centering
  \includegraphics[width=\columnwidth]{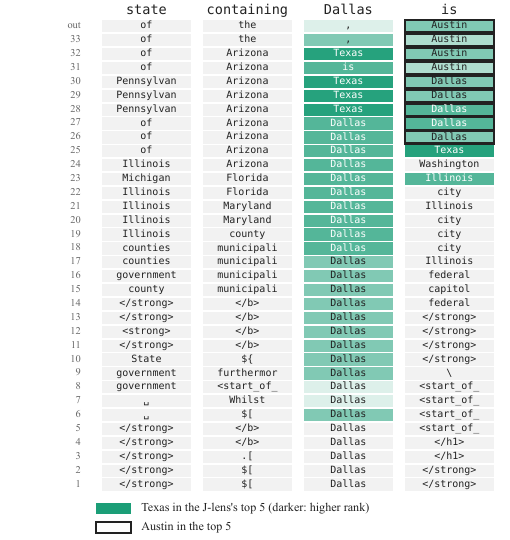}
  \caption{Reading the Dallas question in the viewer (Gemma-3-4B-it): the last four tokens at every layer, each
    cell labeled with the \JL{}'s top word. Shaded: \emph{Texas} is among the top five; outlined: \emph{Austin} is.
    \emph{Texas} is readable at \texttt{Dallas} from layer~6, long before the cell's top word changes, and reaches
    the last token at layer~23, three layers before \emph{Austin}.}
  \label{fig:twohopreading}
\end{figure}

\begin{figure*}[tb]
  \centering
  \includegraphics[width=\linewidth]{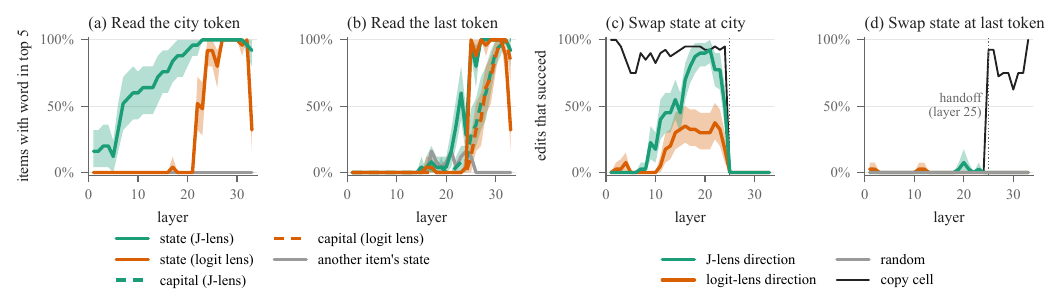}
  \caption{Two-hop recall in Gemma-3-4B-it (``Fact: the capital of the state containing Dallas is'', 25 U.S.\
    items the model answers correctly). (a)~At the city token, the \JL{} reads the state among its top five words
    from about layer~7, the \LL{} from layer~22. As a control, a wrong state (one belonging to another item, such as
    California for Dallas) never enters either lens's top five. (b) At the last token, both lenses read the state before the capital; the J-lens begins to read it a few layers before
    the logit lens does. (c)~Swapping the state for another item's at the city token ($s=4$, 40 pairs)
    gives the new state's capital, nearly as often as the copy cell edit (black line), until the
    handoff. (d)~The same swap at the last token fails, although the state is readable there and copying the cell
    works after the handoff. Bands are bootstrap 95\% CIs.}
  \label{fig:twohop}
\end{figure*}

\textbf{Reading:} The viewer shows the first hop at the city token (\cref{fig:twohopreading}). Highlighting
\emph{Texas} on the Dallas question marks the \texttt{Dallas} cells from layer~6 up, while their top word is still
\emph{Dallas}; at the last token, \emph{Texas} appears at layer~23 and \emph{Austin} only at layer~26. Over all
items, the \JL{} reads the state among its top five words at the city token from a median of layer~7 (U.S.) and 15
(world; \cref{fig:twohop}a); the \LL{} reads it only from layers 22 and 24.

\begin{figure}[H]
  \centering
  \includegraphics[width=\columnwidth]{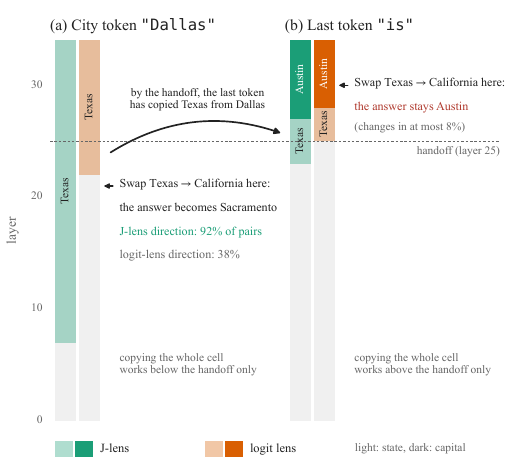}
  \caption{Scenario B in one picture (Gemma-3-4B-it, 34 layers; ``Fact: the capital of the state containing Dallas
    is'', 25 U.S.\ items). Each token has one lane per lens, shaded from the median layer at which the state
    (\emph{Texas}, light) and then the capital (\emph{Austin}, dark) enter the lens's top five. (a)~At the city token
    \texttt{"Dallas"}, the \JL{} reads the first hop from layer~7, well before the \LL{} (layer~22), and swapping
    \emph{Texas} for \emph{California} there makes the model answer \emph{Sacramento}. (b)~By the handoff (layer~25),
    the last token \texttt{"is"} has copied the state from the city token, and both lenses read the state before the
    capital. Swapping the state here leaves the answer \emph{Austin}: the state is readable at this token but not
    used. World items (not shown) read the capital as early as the country, as a memorized shortcut would.}
  \label{fig:twohopmech}
\end{figure}

At the last token, U.S.\ items show the
state before the capital in 92\% of cases under the \JL{} (median layers 23 and 27 of 34; \cref{fig:twohop}b), a
serial pattern consistent with two hops. The \LL{} shows the same order, but reads the state there only from
layer~25. A control state from another item is in the top five for at most 16\% of items at any layer. World items
do not show the pattern: under the \JL{} the capital appears at the same layer as the country or earlier in 77\% of
items, as a memorized city$\to$capital shortcut would. The two lenses thus show different things. The \LL{} reads
what a cell is about to predict as the next token, so it finds the state at \texttt{Dallas} only late, and the
intermediate step is easy to miss. The \JL{} also reads what the cell passes on to later tokens, so it shows the
intermediate step, \emph{Texas}, while the model is still working toward the answer.

\textbf{Editing:} Swapping the state at the city token (\emph{Texas}$\to$\emph{California} at \texttt{Dallas},
$s=4$; \cref{fig:twohopviewer}a--b) with the \JL{} direction gives the new capital in 92\% of 40 pairs (U.S.,
layer~21, CI 82--100\%) and 98\% (world, layer~18, CI 92--100\%), as often as copying the whole cell from the Oakland
prompt (\cref{fig:twohop}c). The \LL{} direction reaches 38\% and 25\%, and swapping the \emph{city} itself
(\texttt{Dallas}$\to$\texttt{Oakland}) only 65\% and 85\%: the city token's state is the variable the model reads.
Random edits never succeeded in our trials.

\textbf{Being readable does not mean that the cell is used.} \emph{Texas} is readable at two places: the city token and, from
layer~23, the last token (\cref{fig:twohopreading}). To see which one the model uses, we can take the same question
about another city (Oakland) and copy one cell from that run into the Dallas run. Copying the \emph{city token}'s
cell changes the answer if done at layer~24 or below, and has no effect from layer~25. Copying the \emph{last
token}'s cell does the opposite: no effect up to layer~24, and it changes the answer from layer~25. So until
layer~24 the answer depends more on the city token, and from layer~25 on the last token. Swapping the state at the last token, where the \JL{} reads \emph{Texas} after this move,
still does not redirect the answer: it gives the new state's capital for at most 8\% (U.S.) and 22\% (world) of
pairs (\cref{fig:twohop}d). On the Dallas question it makes the model say \emph{California}, the state, rather than
\emph{Sacramento} (\cref{fig:twohopviewer}c--d). A static map of lens readings would show \emph{Texas} at both
tokens. Editing shows that the model computes the capital from the copy at the city token, not from the one at the last token.

\section{Scenario C: Answers Before Their Reasoning}
\label{sec:cot}

\begin{figure}[tb]
  \centering
  \includegraphics[width=\columnwidth]{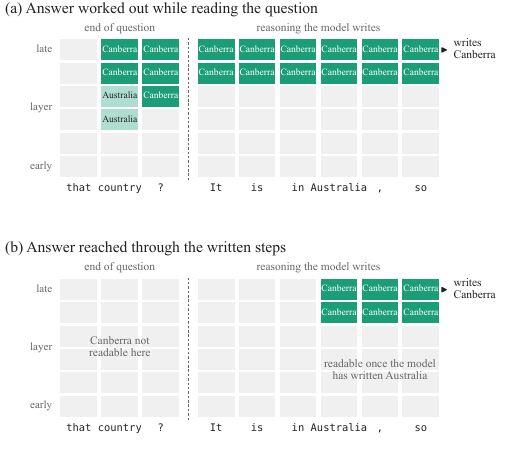}
  \caption{The two patterns we look for (a schematic, not data), for the prompt ``The Sydney Opera House is in a
    country. What is the capital of that country? Think step by step.'' Each cell can only use the tokens to its left.
    (a)~If the model resolves both hops in its layers while reading the question, a lens reads \emph{Australia} and
    then \emph{Canberra} in the question's cells, before any reasoning is written. (b)~If the model needs its written
    steps, \emph{Canberra} becomes readable only after it has written \emph{Australia}. \Cref{fig:cot} counts how
    often each case occurs.}
  \label{fig:cotschematic}
\end{figure}

When a model reasons step by step, does it reach the answer through the steps it writes, or has it already worked
the answer out from the question before writing them? Each cell can only depend on the tokens before it, so the cells
of the question cannot see the reasoning the model writes after it. But if the model works out the answer from the
question alone, a lens can read the answer word in those cells, before any reasoning exists. The viewer lets us check
this one question at a time: let the model answer, highlight its final answer, and look for that word in the cells of
the question, before the first reasoning token (\cref{fig:cotschematic}).

\begin{table}[H]
  \caption{The four task types of the chain-of-thought study, 100 questions each, and how many Gemma-3-4B-it
    answers correctly.}
  \label{tab:cottasks}
  \scriptsize
  \centering
  \begin{tabular}{@{}lp{0.52\columnwidth}r@{}}
    \toprule
    Task type & Questions & Correct \\
    \midrule
    Retrieval & One fact: science questions (SciQ) and country capitals & 80 \\
    Multi-hop & Two facts: a fact about the country where a celebrity was born (its capital, currency, \dots) & 44 \\
    Serial    & Several steps in a row: deduction (ProofWriter) and state tracking (bAbI) & 73 \\
    Ordering  & Arrange several objects from clues: logical deduction (BIG-Bench Hard) & 61 \\
    \midrule
    Total     & & 258 \\
    \bottomrule
  \end{tabular}
\end{table}

For example, we ask ``The Sydney Opera House is in a country. What is the capital of that country? Think step by
step.'' The \JL{} ranks \emph{Canberra} first at the question mark from layer~25 of 34, and the \LL{} from
layer~27, long before the model writes \emph{Australia} in its reasoning: the case of \cref{fig:cotschematic}a. We
chose a landmark outside the capital, so that the lens cannot be reading the landmark's own city. To see how general this is, we ran the same check on
400 questions answered step by step by Gemma-3-4B-it, 100 in each of four task types (\cref{tab:cottasks}). We keep
the 258 questions the model answers correctly, and call an answer readable before the reasoning if a lens ranks it
first at any layer of any token before the first reasoning token; we also report top-five rates.  As a control, each question also has a plausible wrong answer of the same kind, the \emph{distractor}: SciQ uses its own distractor options for science (e.g.\ \emph{created} for \emph{destroyed}), another capital or country is used for the fact and multi-hop tasks (e.g.\ \emph{Lisbon} for \emph{Madrid}), one of the remaining truth values is used for deduction (e.g.\ \emph{False} for \emph{True}), and another named entity from the prompt is used for state-tracking and ordering (e.g.\ \emph{hatchback} for \emph{convertible}). When a question yields several distractors, we score the one a lens ranks highest, the strongest control. We count how often a lens ranks this distractor first in the same way. If
the true answer is readable much more often than the wrong one, the lens is reading the model's answer, not just the
kind of word the question asks for.

\begin{figure}[tb]
  \centering
  \includegraphics[width=\columnwidth]{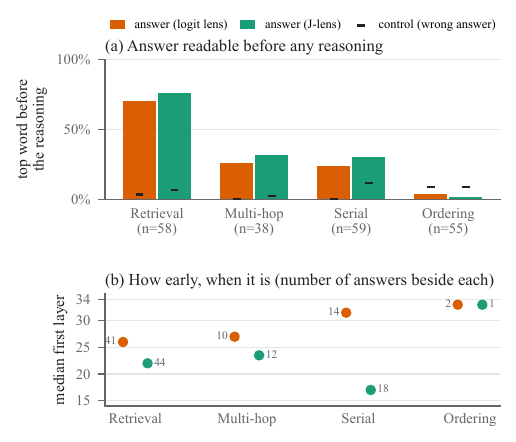}
  \caption{Is the final answer readable before the reasoning? Gemma-3-4B-it, correctly answered questions out of 100
    per task type. (a)~Share of answers that a lens ranks first at some cell before the first reasoning token; the
    black ticks are a plausible wrong answer (control). (b)~For those answers, the median first layer at which the
    lens reads it, with the number of answers beside each point.}
  \label{fig:cot}
\end{figure}

\textbf{The answer is often readable before the steps.} For retrieval, the answer is the top word before any
reasoning for 89\% (\LL) and 92\% (\JL) of correct answers, and among the top five for 96\% and 96\%; the wrong
answer is the top word for at most 5\% (\cref{fig:cot}a). Multi-hop and serial questions are readable less often, but
still well above the control: 39\% and 43\% for multi-hop and 53\% and 63\% for serial, against at most 5\%.
So for between two fifths and nine in ten of these questions, the reasoning the model writes comes after an answer that is
already in its residual stream.

\textbf{The \JL{} reads the answer earlier.} When the answer is readable, the \JL{} reads it at the same layer as
the \LL{} or a layer or two earlier in every task type (\cref{fig:cot}b): a median of layer~24 against~25 for
retrieval, 24 against~26 for multi-hop, and 24 against~25 for serial. The effect is consistent in direction but
small, one to two layers.

\textbf{Ordering is worked out in the reasoning steps.} For ordering puzzles, neither lens reads the answer before the
reasoning: it is the top word for 7\% (\LL) and 5\% (\JL) of correct answers, no more often than the wrong answer
(16--20\%), and in the top five for only 10\% and 16\%. Here the model appears to reach the answer while writing the steps,
not before them.

\section{Discussion}
\label{sec:discussion}
\;
\textbf{Readings gave us hypotheses to test.}
In each scenario, a lens reading provided a glimpse of the model's
thought process, and an edit decided
whether that mechanism held. Sometimes the edit confirmed the mechanism: \emph{Texas} is readable at \texttt{Dallas} from layer~7, and
swapping it there redirects the answer 92--98\% of the time. Sometimes the edit contradicted it: \emph{Texas} is just
as readable at the last token after the handoff, but swapping it after the handoff layer almost never changes the answer.
A heatmap shows the same word in both places and cannot tell which one the model uses. We built MicroLens
because it lets us causally probe how a language model reasons and where its answers form.

\textbf{The \JL{} and \LL es are complementary.} The \JL{} reads ideas earlier than the \LL: the intermediate state in
two-hop questions (layer~7 against 22), and answers before their reasoning in every task type, by one to two layers. But the lens that reads an idea earliest is not the one that writes an output best. To set what
a token outputs, the \LL{} direction works from layers 6--8 and the \JL{} direction only from 22--30; to change what
later tokens read, the \JL{} direction is the one that works. A viewer should therefore offer both lenses for reading,
and choose the direction for each edit by its purpose.

\textbf{When do edits work?} We found two limits on edits recurring across the tested scenarios in Gemma-3-4B-it.
First, edits worked only before the handoff. Once later tokens had read the word from its cell, editing that cell no
longer changed the answer. Copying the word's cell from a prompt that names a different entity, one layer at a time,
found the handoff: the copy changed the answer below the handoff and had no effect from it onward
(\cref{fig:handoff}). Second, a \JL{} swap worked only when the \JL{} read the word in its own cell. For names, this
was true from the early layers. For words with many senses, this only happened after the handoff, when it was too late
to edit.

\section{Conclusion}

Lens heatmaps give intuition about what a language model computes at a given cell, but without an easy way to edit
each cell and see what changes, we cannot clearly trace how the model uses that information to reach a final answer.

MicroLens puts reading and editing on the same grid, so that a user can edit any cell right where the reading was
made and watch how the answer changes. With it,
we found that a \JL{} swap works only where
the \JL{} can read the word. We also located where a two-hop question's intermediate entity is used and where it is merely
readable. And we found that, except on ordering puzzles, the model's answer is often readable before it writes any
reasoning, a layer or two earlier through the \JL{} than the \LL.

\section{Limitations}
\begin{itemize}[nosep, leftmargin=*]
  \item \emph{One model:} Gemma-3-4B-it only; a Gemma-3-1B pilot on country--capital shows the same split
        (\LL{} push from layer~6 of 26, \JL{} from 21; \JL{} swap 90\% vs.\ 23\%).
  \item \emph{Narrow inputs:} one prompt format per task, single-token words, correct items only, 25--100 items per
        condition.
  \item \emph{Averaged directions:} the \JL{} reflects a word's average meaning, so its swaps reach 62--79\% of copying
        the cell on facts and fail on many-sense words.
  \item \emph{Approximate trace:} an attention rollout that ignores MLPs; a guide to where to edit, not proof.
  \item \emph{No edits in Scenario~C:} the answer is present before the reasoning, but we do not show that it is used.
  \item \emph{No external users:} our own usage scenarios backed by sweeps, not a user study.
\end{itemize}
\textbf{Future work:} The viewer makes the missing test for Scenario~C easy to run: swap the answer at the
question's cells and see whether the reasoning, or only the final answer, changes. Other next steps are a user
study with interpretability researchers, more model families, and edit directions that adapt to the prompt rather
than averaging over contexts.

\acknowledgments{
We used Claude (Anthropic) as a coding and writing assistant. It helped implement parts of the viewer, run some of the experiment scripts, produce and reformat figures, and edit the text for clarity. The authors designed the study and experiments,
checked every result against the experiment outputs, and take full responsibility for the content of this paper.
}

\section*{Conflict of Interest}
Fangyi Zhu holds stock in Alphabet Inc., the parent company of Google, which developed the Gemma model evaluated in this study.
Fangyi Zhu was employed by Alphabet within the past 36 months.
The remaining authors declare no competing interests.

\bibliographystyle{abbrv-doi-hyperref}
\bibliography{refs}

\end{document}